\documentclass[conference]{IEEEtran}
\IEEEoverridecommandlockouts
\usepackage{cite}
\usepackage{amsmath,amssymb,amsfonts}
\usepackage{algorithmic}
\usepackage{graphicx}
\usepackage{textcomp}
\usepackage{xcolor}
\usepackage{mathtools}
\def\BibTeX{{\rm B\kern-.05em{\sc i\kern-.025em b}\kern-.08em
    T\kern-.1667em\lower.7ex\hbox{E}\kern-.125emX}}

\usepackage{todonotes}

\usepackage{booktabs} 
\usepackage{hyperref} 
\usepackage[linesnumbered,ruled,vlined]{algorithm2e}
\DontPrintSemicolon
\DeclareMathOperator{\attack}{attack}
\DeclareMathOperator{\cost}{cost}

\begin{document}
     
\title{Adversarial Training for a Continuous Robustness Control Problem in Power Systems
}

\author{\IEEEauthorblockN{Loïc Omnes}
\IEEEauthorblockA{\textit{AI Lab} \\
\textit{RTE}\\
Paris, France \\
loic.omnes@free.fr}
\and
\IEEEauthorblockN{Antoine Marot}
\IEEEauthorblockA{\textit{AI Lab} \\
\textit{RTE}\\
Paris, France \\
antoine.marot@rte-france.com}
\and
\IEEEauthorblockN{Benjamin Donnot}
\IEEEauthorblockA{\textit{AI Lab} \\
\textit{RTE}\\
Paris, France \\
benjamin.donnot@rte-france.com}

}

\IEEEoverridecommandlockouts
\IEEEpubid{\makebox[\columnwidth]{978-1-6654-3597-0/21/\$31.00~\copyright2021 IEEE \hfill} \hspace{\columnsep}\makebox[\columnwidth]{ }}
\maketitle
\IEEEpubidadjcol

\begin{abstract}
We propose a new adversarial training approach for injecting robustness when designing controllers for upcoming cyber-physical power systems. Previous approaches relying deeply on simulations are not able to cope with the rising complexity and are too costly when used online in terms of computation budget. In comparison, our method proves to be computationally efficient online while displaying useful robustness properties. To do so we model an adversarial framework, propose the implementation of a fixed opponent policy and test it on a L2RPN (Learning to Run a Power Network) environment. This environment is a synthetic but realistic modeling of a cyber-physical system accounting for one third of the IEEE 118 grid. Using adversarial testing, we analyze the results of submitted trained agents from the robustness track of the L2RPN competition. We then further assess the performance of these agents in regards to the continuous N-1 problem through tailored evaluation metrics. We discover that some agents trained in an adversarial way demonstrate interesting preventive behaviors in that regard, which we discuss.

\end{abstract}


\section{Introduction}

Power systems have always been critical systems in which robustness is crucial \cite{bevrani2009robust}. For the last decades, developed power grids have tried to remain robust to unexpected events following a simple guiding “N-1” redundancy principle: the system  should  keep  operating  in  nominal  conditions  despite  the  loss  of  any  single  asset such as lines or power plants at any time.  Operators hence anticipated through forecasts and simulations the effect of having any line disconnected in the hours to come.  This very fundamental  property  has  been  the  root  of the success of power  systems : to distribute  electricity reliably throughout  the  grid at  all  times. However, these aging systems are moving closer to their limits and this redundancy constraint is no longer easy to satisfy due to the ever increasing complexity. In addition, climate change can potentially lead to more frequent and extreme weather events which also have to  be  accounted  for. As a consequence, the time resolution operators are considering is moving towards sampling and refresh rates of 15 minutes instead of 3 hours previously. Because of constraints over the computation time budget in real-time operations, not every contingency can be studied on these higher resolution forecasting horizon. There is therefore a need for new methods to be robust to extreme events while having the capacity to operate in real time even at larger scales. That need has led to a trend toward cyber-physical systems with innovative behaviors \cite{allgoewer2019position}.

Some methods have successfully addressed this problem before (worst case methods \cite{capitanescu2011day}, robust control) but they have troubles scaling up to continuous time horizon and adapting to cyber considerations. It is therefore essential to find new ways of coming up with robust strategies that are capable of coping with the new complexity in the power systems.

Machine Learning seems to be a promising approach to solve this issue. Indeed, learning a certain behavior beforehand could allow to constantly use that knowledge instead of making use of the computation budget to simulate events and thus gain significant improvements in online computation times as shown by \cite{donnot2018optimization}, possibly solving the scaling issue. In particular, it could be possible to predict power flows using Deep Learning. Other kinds of Machine Learning methods look particularly suited to the problem, such as Reinforcement Learning methods \cite{dulac-arnold_challenges_2019}\cite{dalal2016hierarchical}. These methods let agents learn through experience, possibly solving extremely complex problems \cite{silver2017mastering}. Reinforcement Learning techniques are also adapted for finding robust policies \cite{morimoto_robust_2001}. Moreover, once the modeling framework has been built, environments for different networks in the world can be created easily and quickly, allowing for the models to be used for the different existing networks. There is a wide access to experience data in power systems that could allow for Reinforcement Learning methods to thrive.  

Thus, control is now a central area of power systems. Recent research recommends worst case methods, adversarial testing and safe learning \cite{dobbe2020learning}. Following these guidelines, we seek to achieve safe learning through an adversarial training approach. Furthermore, we make use of the L2RPN (Learning to Run a Power Network) competitions environments \cite{marot2020learning} to evaluate the submitted agents using adversarial testing. In this paper we present the opponent that we considered, we propose to characterize in a more exhaustive way the robustness and we evaluate the relevance of adversarial training as a means of building robust policies.

\section{Formalization of the N-1 problem}

The initial objective for the robustness competition is that the agent should be able to solve the continuous N-1 problem over time, simply referred as the N-1 problem.
The N-1 problem is a well-known problem in power systems \cite{abulwafasecurity}.

\subsection{Definition of N-1 time-continuous problem}

At a given time, the instantaneous N-1 problem is to keep operating the network despite the loss of any element of the electrical network at that moment. That is, at that instant, one must be robust to the loss of any electrical network element.

Therefore, the continuous N-1 problem is the following: to be able to keep operating the network over a time horizon, despite the untimely loss of an element of the electrical network.
This implies a notion of robustness because it is necessary to prepare in advance for the eventual loss of an element, hence the name N-1.
Specifically, one must not only be prepared for the loss of an element but also have the grid be in a configuration that is acceptable for the hours to come when a contingency happens. 


Our goal is on the one hand to create a framework that models and implements the N-1 problem and that provides a means to evaluate the robustness of an agent, and on the other hand to propose agents that can respond to it, i.e. agents that are robust to the loss of an element of the power grid.

\subsection{Metrics to evaluate the N-1}

We first look for a meaningful metric in regards to the N-1 problem. The objective is to find a single quantity that is capable of assessing an agent's performance relatively to the N-1 problem at some time step $t$. We call it the N-1 reward, $R_t^{eval}$.
Other studies have aimed at evaluating the N-1 criterion before, typically through an evaluation reward equally considering the different line disconnections \cite{dalal2016hierarchical} :

\begin{equation}
R_t^{eval}=\Sigma_{i=1}^{n_{lines}}S_{i,t}
\end{equation}

Where :
\begin{itemize}
\item $n_{lines}$ is the number of attackable lines.
\item $S_{i,t}$ is a stability score to be determined for when the disconnection of the line $i$ is simulated at time step $t$. \\
In the previous example \cite{dalal2016hierarchical}, $S_{i,t}$ equals 1 if the grid is safe, else 0. \\
\end{itemize}

Such metrics account for the robustness to the different possible line disconnections uniformly. At the contrary, worst case approaches only consider the worst possible line disconnection at some given time \cite{mankowitz_robust_2020}\cite{dulac-arnold_challenges_2019}\cite{xiao_adversarial_training}, using :

\begin{equation}
R_t^{eval}=min_{i \in [1, n_{lines}]}S_{i,t}
\end{equation}

We propose another approach in-between those two that accounts for all the different possible line disconnections while putting an emphasis on the worst ones. \\

We remind that the N-1 problem is to be robust to the untimely disconnection of one of the attackable lines.

Therefore to evaluate it, we choose to simulate exhaustively at each step the disconnection of each of those attackable lines and check whether any overflow appeared in the power grid. That way, we know if the agent was prepared for such a disconnection.
Note that those simulations can be computationally expensive but are only required for the evaluation of the agents and not when simply running them. \\

We then put together these values $S_{i,t}$ into the following formula for the N-1 reward $R_t^{eval}$ :

\begin{equation}
R_t^{eval}=\Sigma_{i=1}^{n_{lines}}w_{\phi_t^{-1}(i)}S_{i,t}
\end{equation}

Where :
\begin{itemize}
\item We choose that the stability score $S_{i,t}$ equals 1 if no overflows occurred in the grid after the disconnection was simulated, else 0.
\item $w_j$ are weights given to the stability scores of the disconnections and will be discussed further down. \\
At a given time step $t$, the highest weights are always given to the lowest scores $S_{i,t}$, hence the permutation $\phi_t$ to reorder the weights to the stability scores. \\
This is done to have a behavior close to the one of worst case evaluation, while still considering all line disconnections instead of only the worst one. \\
\end{itemize}

Through this evaluation reward, we consider all of the different possible line disconnections while assigning higher weights to the most dangerous ones. \\

As for the weights, we choose the following :
\begin{equation}
w_j=exp(-\lambda \frac{j-1}{n_{lines}-1})
\end{equation}

The weights are exponentially decreasing and $w_0=1$, where $\lambda$ is a parameter to be determined and which balances the emphasis to be put on the worst disconnections. Tuning that parameter effectively allows us to get closer to either the worst case metrics or the uniform ones. \\

We choose a value for $\lambda$ such that 95\% of the weights are contained in the 20\% worst line disconnections, i.e. with lowest stability scores $S_{i,t}$ at a given time. This gives strong importance to the worst line disconnections while being soft enough so that the values for the metric can be easily and meaningfully compared across different agents and scenarios.

\section{Modeling \& adversarial approach implementation}

\subsection{Opponent modeling}

At scale, it is impossible to simulate all N-1 disconnections online over a long time horizon because of the computation burden.
Therefore, we propose a new adversarial approach that lets a controller learn offline to be robust to those disconnections without having to simulate them online. \\

Our adversarial approach is inspired from the classical RL (Reinforcement Learning) framework \cite{sutton1998introduction}, where an agent and an environment interact with each other sequentially.

Here we introduce a modified version of this framework where we add a new adversarial actor, the opponent. The opponent acts in parallel to the agent and in a similar way, picking an action at each time step according to some observation for the current state that it received from the environment. The purpose of the opponent will be to trigger untimely adversarial powerline disconnections that the agent then has to cope with. This aims at forcing the agent to acquire some sort of robustness to these disconnections, the objective being that it can then solve the N-1 problem. \\

The observations for the agent mostly contain information concerning power flows, productions, consumptions and the topology of the grid. The actions that are available to the agent are topological modifications such as connecting or disconnecting powerlines and splitting or merging nodes in substations, as well as redispatching actions on productions. Please refer to \cite{kelly2020reinforcement} for more details.

The opponent has access to the same observation as the agent and can choose to either disconnect a line or do nothing.

\begin{figure}[htbp]
\centerline{\includegraphics[width=9cm]{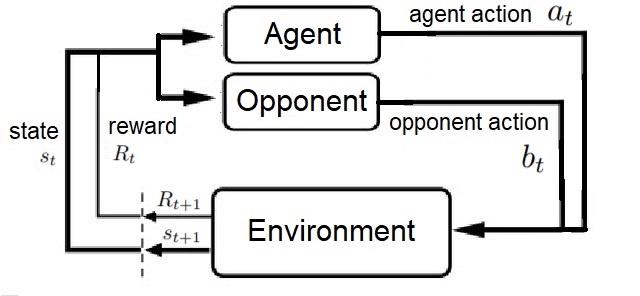}}
\vspace{-4mm}
\caption{Considered adversarial RL framework.}
\end{figure}

\begin{figure}
\centerline{\includegraphics[width=8cm]{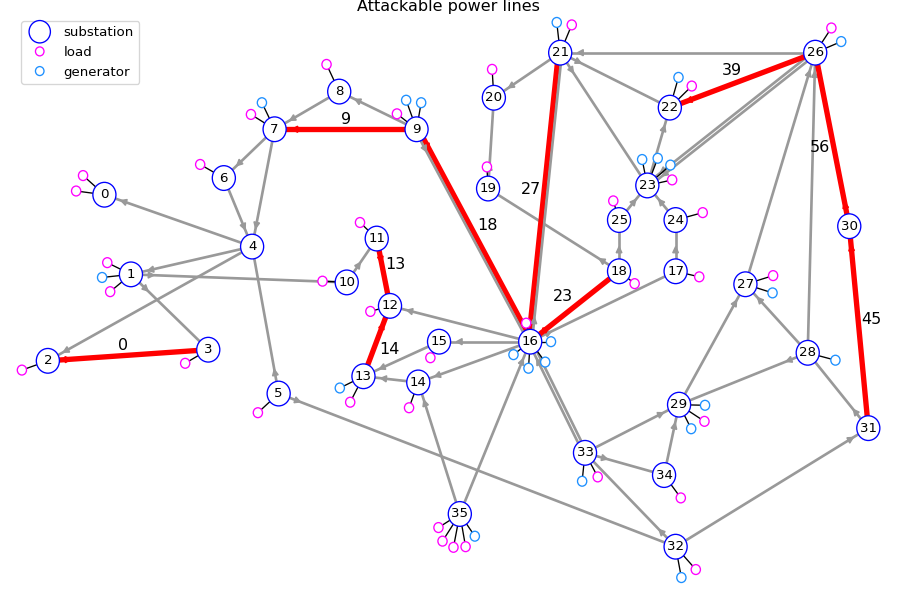}}
\vspace{-3mm}
\caption{10 Attackable powerlines highlighted.}
\label{fig_attackable}
\vspace{-4mm}
\end{figure}

Furthermore, so that the difficulty of the competition is not too high, we chose that the opponent should only be able to disconnect the powerlines that are subject to maintenance. This would make the agents less confused since they already had to face their disconnections at maintenance times. These powerlines are shown in red in figure \ref{fig_attackable}. \\

The agent and the opponent hence have opposite goals that lead to joint equations for their optimal policies :

\begin{equation}
\pi^*=argmax_{\pi \in \Pi} E[\Sigma_{t=0}^{t_{final}}R_t | \pi, \psi^*]
\end{equation}
\begin{equation}
\psi^*=argmin_{\psi \in \Psi} E[\Sigma_{t=0}^{t_{final}}R_t | \pi^*, \psi]
\end{equation}

Where :
\begin{itemize}
\item $\pi$ and $\psi$ are the agent and opponent policies respectively.
\item $\Pi$ and $\Psi$ are the agent and opponent policy domains respectively, which may be different to the total spaces due to constraints over the policies.
\item $R_t$ is the reward received by the agent at time step $t$.
\item $t_{final}$ is the total number of time steps for the scenario.
\end{itemize}

Eventually, an agent robust to the optimal opponent policy $\psi^*$ is also robust to all opponent policies for the N-1 problem.

\subsection{Considering the class of fixed policy opponents for the L2RPN competition}

However, we need to choose a fixed opponent policy for the competition beforehand. Thus, during the competition, only the agent policy is going to change and improve relatively to that opponent policy. This policy will likely be different from the opponent optimal policy $\psi^*$. \\

Then comes the choice of an opponent policy. We consider that it actually has two distinct objectives :
\begin{itemize}
\item His attacks must be as dangerous as possible for the agent as stated before.
\item His attacks must be fairly uniformly distributed among the powerlines.
\end{itemize}

We choose to add that second goal in order to be consistent with reality and to not be too predictable for the participants. \\

Those two goals can easily go in opposite directions since enforcing a uniform distribution acts as a constraint for the effectiveness objective. It is necessary to find a compromise between those two aspects.

We integrate a second term into the opponent's optimal policy equation to reflect that second end of the opponent :
%
\begin{multline}
\psi^*=argmin_{\psi \in \Psi} E[(\Sigma_{t=0}^{t_{final}}R_t+\mu .\delta_{attack})| \pi^*, \psi]
\\ \text{with } \delta_{attack}=max_{i, j \in [1, n_{lines}]^2}|n_i - n_j|
\label{eq:opp_final}
\end{multline}

Where :
\begin{itemize}
\item $n_i$ is total number of times the opponent has attacked the powerline $i$ during the scenario.
\item $\mu$ is a parameter balancing the relative importance between the two terms.
\item $n_{lines}$ is the number of attackable lines in the power grid.
\end{itemize}

The second term of the equation states that the $n_i$ should be as close as possible, meaning that the attack distribution should be as balanced as possible. \\

We try to come up with an opponent for the competition that is close to $\psi^*$. Here are some possible opponents examples :
\begin{itemize}
\item Baseline : DoNothingOpponent. This opponent does nothing, i.e. never attacks. It is useful as a baseline for comparison with other opponents.
\item Proposal : \textbf{WeightedRandomOpponent}. This is the opponent that we bring forward here and that we chose for the competition. It will be detailed in the next subsection.
\item Trained opponent : this is a possibility that we did not explore because it would require having reliable and diverse agents to train against.
\end{itemize}

Before diving into the opponent that we propose, it is important to talk about the constraints that we chose for the opponent.
Indeed, it is necessary for the opponent to have constraints to prevent it from freely disconnecting a line at each time step, which would result in a problem that would be too far from reality and also probably could not be solved, hence not really interesting and useful to study. \\

We choose the following constraints for the opponent :
\begin{itemize}
\item An attack consists in the disconnection of $n_{attack}$ powerlines at a time. 
\item An attack must last for a duration $d_{attack}$. During that time, the powerline cannot be reconnected.
\item It is restricted to a maximum frequency of attacks. It can only attack once in the time period $T_{attack}$. \\
We note $\forall k \in \mathbb{N}, T_k = [k*T_{attack}, (k+1)*T_{attack}[$ the successive attack periods of a scenario.
\item It has a certain budget which increases by a fixed amount at each time step. The opponent can only attack if the cost of the attack is inferior to its current budget. In our case, the cost of any attack is fixed to 1.
\item Only some powerlines in the grid can be attacked by the opponent. It will not be able to disconnect the other ones.
\end{itemize} 

For the competition, we set $n_{attack}=1$, $d_{attack}=4$ hours, $T_{attack}=24$ hours.

\subsection{WeightedRandomOpponent}

For that opponent, for each attack period $T_k$ of a scenario, the time of the attack $t_k$ is uniformly picked at random during that attack period.

The powerline $l_k$ that is attacked at that time is also picked at random among the attackable powerlines, with probabilities $\rho_{i, t} / \alpha_i$ if the attack is made at time step $t$, where $\rho_{i, t}$ is the electric current flowing in the powerline $i$ at time step $t$ relatively to the thermal capacity of the line, and $\alpha_i$ is a normalizing factor equal to the empirical mean value of $\rho_{i, t}$ through time. We note this probability distribution $D_{\rho_t}$.

Formally, for an environment state $s$, the WeightedRandomOpponent policy $\psi_{wro}$ is as follows :
\begin{equation}
\begin{multlined}[t]
\psi_{wro}(s) = \attack(l_k) \text{ if } t(s) \in \{t_k\}_{k \in \mathbb{N}} \text{ else nothing} \\
\text{where } \forall k \in \mathbb{N}, t_k \sim U(T_k) \text{ and } l_k \sim D_{\rho_{t_k}(s)} \\
\text{s.t.} \cost(\attack(l_k)) \leq \text{budget} \\
t_k + d_{attack} \leq t_{k+1} \\
\label{eq:wro_policy}
\vspace{-3mm}
\end{multlined}
\end{equation}
%
%
%

This opponent is within the constraints we mentioned and aims at fulfilling the two objectives of the opponent highlighted in the equation \ref{eq:opp_final}. Indeed, on the one hand it favors the most dangerous attacks since the most loaded powerlines have a higher chance of being attacked and are more likely to put the agent in danger when disconnected. On the other hand, the normalizing factors $\alpha_i$ ensure a balanced distribution of attacks between the different powerlines.

At the same time, the behavior of the opponent is randomized concerning both the time of the attack and the line to attack. This is so that it is not easily predictable for the participants of the competition.

\section{Experiments \& results}

\subsection{Competition results}

The results and winners of the WCCI (World Congress on Computational Intelligence) and NeurIPS (Neural Information Processing Systems) competitions can be checked \href{https://l2rpn.chalearn.org/competitions}{\underline{online}}. The competition score is based on the cumulative network operational cost and normalized to the range [-100, 100] where the score is -100 for an initial blackout, 0 for when the agent does nothing (do\_nothing agent), 80 for when the scenario is completed and up to 100 depending on the operational cost optimization. Refer to \cite{marot2021l2rpn} for more details. \\

For our experiments, as a first step we picked four of the top submitted agents from the WCCI and NeurIPS robustness competitions \cite{marot2020l2rpn} and ran them against those two competitions along with two of our baseline agents. The data we used was generated with \href{https://github.com/BDonnot/ChroniX2Grid}{\underline{Chronix2Grid}}. The results are shown in table \ref{tab:pres_results}.
\begin{table}
  \caption{Agents results on the WCCI and NeurIPS competitions}
  \centering
  \begin{tabular}{llll}
    \toprule
    \multicolumn{2}{c}{} & \multicolumn{2}{c}{Results} \\
    \cmidrule{3-4}
    Agent     & From competition & WCCI   & NeurIPS \\
    \midrule
    rl\_agnet         & NeurIPS  & 71.21  & 61.05     \\
    binbinchen        & NeurIPS  & 63.97  & 52.42     \\
    lujixiang         & NeurIPS  & 73.73  & 45.00     \\
    zenghsh3          & WCCI     & 58.21  & 19.10     \\
    reco\_powerline   & Baseline & 25.75  & 10.76     \\
    do\_nothing       & Baseline & 0.00   & 0.00      \\
    \bottomrule
  \end{tabular}
  \label{tab:pres_results}
\vspace{-5mm}
\end{table}
The only major difference between the two competitions lies in the fact that the NeurIPS competition uses an opponent to threaten the agents, while the WCCI competition does not. Therefore, the superiority of the NeurIPS agents over one of the top WCCI contestants, zenghsh3, could suggest an effectiveness of the adversarial training method to which the NeurIPS agents were confronted.

Not only are these agents stronger on the NeurIPS Robustness competition, they also outperform the WCCI agent on the WCCI competition. 
Plus, while the scores for the NeurIPS agents are only a little bit lower on the NeurIPS competition, which is expected since this competition uses an opponent and is thus harder to deal with, the performance of zenghsh3 drops much more dramatically when confronted to the opponent.

It should be noted that contrarily to the other two NeurIPS agents, lujixiang's score also drops more significantly on the NeurIPS competition but this was also expected since we noted that it was already less robust on the WCCI competition compared to the other agents, failing at one scenario, despite obtaining a very high score thanks to better energy loss management. As a whole, these results seem to indicate not only a better performance in general of the agents which have used adversarial training, but also a stronger robustness to line disconnections. However, since we only have one strong agent to study from the WCCI competition, we cannot fully conclude on that matter given only these data. \\

In this part we have evaluated the agents robustness through their competition scores, in particular on the NeurIPS competition, which is actually adversarial testing.

Next, we come up next with different metrics and evaluations that fit better the N-1 problem in order to have a more meaningful measure of what we want to achieve as well as a deeper insight on the agents behaviors.

\subsection{Evaluation based on the N-1 criterion}

It is important to note that we conduct a preventive robustness study here but do not evaluate the agent's corrective behavior since that metric is based on simulated disconnections and is measured before the agent has a chance to respond with an action. Thus, it evaluates how well the agent is prepared but not how well the agent could correct an issue.

We then run exhaustive experiments for some of the best agents from the NeurIPS competition and some of our baselines, on the 24 NeurIPS competition test scenarios, and measure their performance based on the N-1 reward $R_t^{eval}$ averaged over these scenarios.

\begin{figure}
\hspace{2mm}
\centerline{\includegraphics[width=9cm]{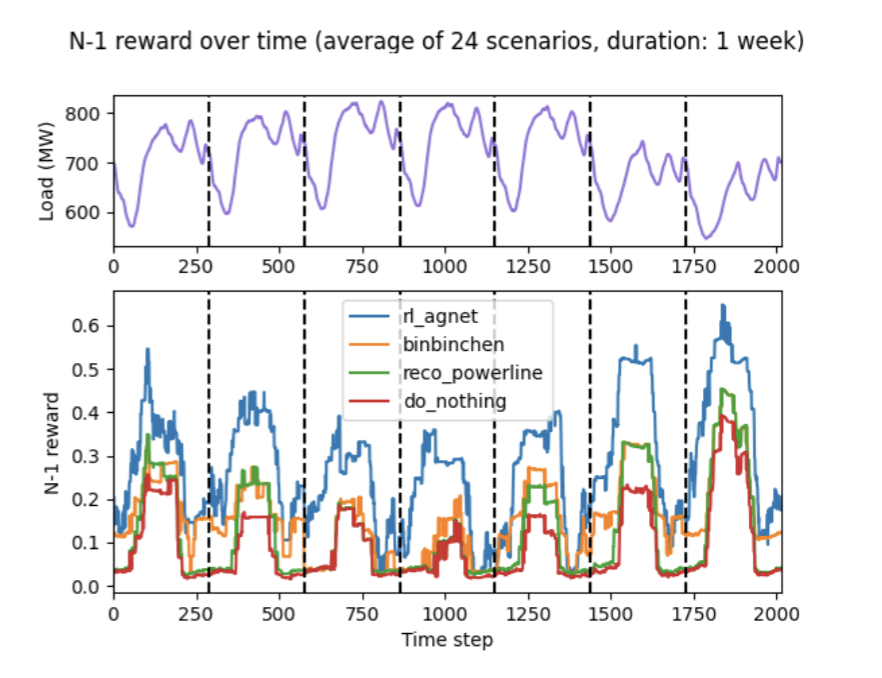}}
\vspace{-9mm}
\caption{Mean N-1 reward through time for one-week long scenarios.}
\vspace{-3mm}
\label{fig_series}
\end{figure}

During all the following experiments, no opponent has been used since the N-1 problem evaluation is already contained in the N-1 reward.
These scenarios each last one week and go from Monday to Sunday. The results are shown in figure \ref{fig_series}. Through that experiment, we want to evaluate if an agent actually more robust to attacks is ultimately more robust to the N-1 continuous problem.

The N-1 rewards and the load in the power grid are strongly linked. For each agent, the N-1 rewards tend to increase when the load is low and to decrease when the load is high. Indeed, the correlation between the load and the smoothed derivative of the N-1 rewards is as significant as -0.5. This is expected as a higher load means more loaded lines and thus a higher probability to see congestion appearing on the grid, hence a lower reward. \\

\begin{figure}
\hspace{5mm}
\centerline{\includegraphics[width=9cm]{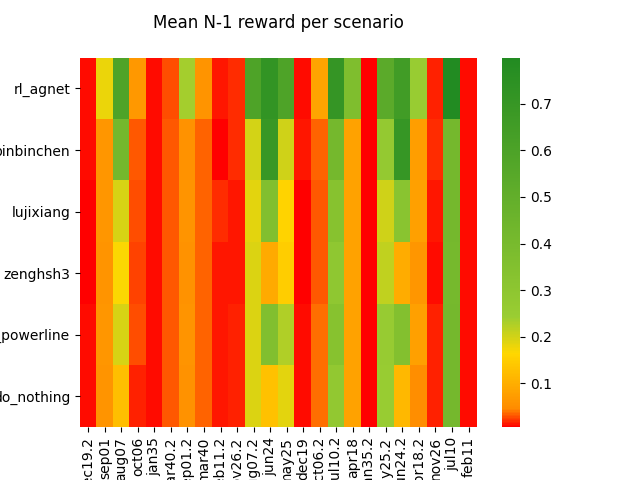}}
\vspace{-6mm}
\caption{Mean N-1 reward for each of the 24 NeurIPS test scenarios.}
\label{fig_heatmap}
\vspace{-5mm}
\end{figure}

A clear gap is to be observed at all times between rl\_agnet and the other agents. Indeed, as already indicated in the competition scores, that agent does much better than the other ones when evaluated with the N-1 reward, reinforcing the idea that this agent acquired a strong notion of robustness to the disconnection of the powerlines. Next come reco\_powerline and binbinchen. Except sometimes at night, we don't notice any substantial gain from binbinchen agent there: we can wonder why given its good score in the competition. To study more carefully this observation, we also choose to inspect the mean N-1 rewards for each scenario individually in order to have a clearer view of the agents performance (see figure \ref{fig_heatmap}).

The rl\_agnet agent has again the highest mean N-1 reward on almost all scenarios, accounting once more for his higher robustness. The other NeurIPS agents, binbinchen and lujixiang, come next along with the baseline reco\_powerline. Their performance as evaluated by the N-1 reward is rather disappointing given that they both reached high scores in the competition. We believe this can be explained by the fact that we only evaluate the preventive capacity of the agents and they most probably rely strongly on their corrective capacities. This will be the object of future works.

It can be seen that the WCCI agent zenghsh3 perform slightly worse than all other agents on almost all scenarios, suggesting again that the agents which have used adversarial trained have effectively acquired a stronger robustness to powerline disconnections.

Yet, there are still several difficult scenarios during which the loads are rather high and in which no agent could really conserve a durable robustness. \\

Lastly, we inspect the agents robustness to each line disconnection separately to see if a robustness has been acquired uniformly throughout the grid as desired, or if the management of some of the powerlines has been prioritized.

\begin{figure}
\centerline{\includegraphics[width=9cm]{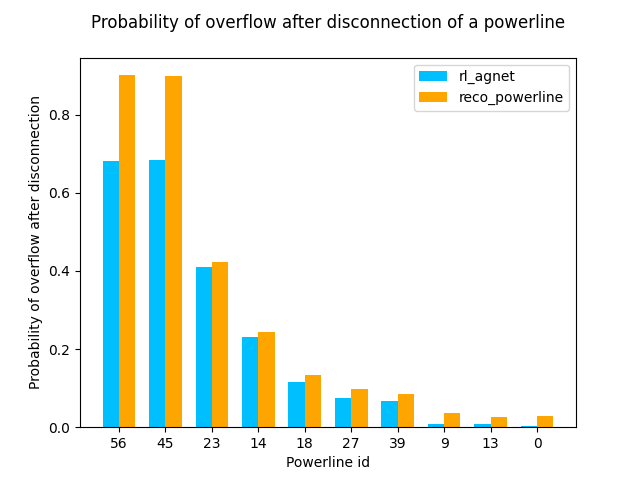}}
\vspace{-3mm}
\caption{Empirical probabilities of overflow after disconnection of a powerline, for each attackable powerline. The confidence intervals are negligeable given the high number of observations (less than $10^{-3}$ wide).}
\label{fig_barplot}
\vspace{-3mm}
\end{figure}

We show the probabilities a disconnection of a line will cause an overflow, averaged over the 24 NeurIPS test scenarios, for rl\_agnet compared to reco\_powerline in Figure \ref{fig_barplot}.

It can be seen that for all the attackable lines in the grid, the probability of a disconnection causing an overflow is lower for the rl\_agnet agent than for reco\_powerline. This suggests that the agent has obtained a robustness to line disconnections not only in some parts of the power grid but in the whole grid, making it a robust and versatile agent. This what we hoped for when designing our opponent and we believe it helped achieve that robustness property. Studying the impact of different opponents on the learnt agent robustness will be the object of future works to draw more define conclusions about the opponent design. 
Adversarial training hence showed to be an appropriate framework for an agent to become robust to the N-1 problem, and this without having to run costly online N-1 simulations to ensure it.

\section{Conclusion}

In this paper, we presented an original approach for learning a controller, i.e. an agent, with desirable robustness properties, in particular according to the N-1 principle. We achieve interesting online computational efficiency and robust performance without online N-1 simulations thanks to our adversarial training. The best agent further displays a preventive behavior in addition to any curative actions, highlighting an advanced robust strategy compared to other agents. There are possible extensions to this paper that we leave as future works such as extending the set of attackable lines, the evaluation of corrective behavior in addition to the preventive one, as well as the design of new opponents with trained policies.

\section*{Acknowledgment}

We thank Camilo Romero and Jean Grizet for making the L2RRPN competitions and environments possible. We thank Isabelle Guyon, Gabriel Dulac and Patrick Panciatici for their valuable insights for framing the problem and guiding us through relevant modelizations.

\bibliographystyle{ieeetr}
\bibliography{adversarial_robustness}

\end{document}